\documentclass[runningheads]{llncs}
\usepackage{graphicx}
\usepackage{tabularx}
\usepackage{multirow}
\usepackage[T1]{fontenc}
\usepackage{amsmath}
\usepackage{amsfonts}
\usepackage{amssymb}
\usepackage{wrapfig}
\usepackage{tikz}
\usepackage{comment}
\usepackage{amsmath,amssymb} % define this before the line numbering.
\usepackage{color}
\usepackage{multirow}
\usepackage{kotex}
\usepackage{adjustbox}
\usepackage{float}
\usepackage{booktabs}  % professional-quality tables
\usepackage{colortbl}

\begin{document}

\title{Brain-Streams: fMRI-to-Image Reconstruction with Multi-modal Guidance}

\author{Jaehoon Joo, Taejin Jeong, Seong Jae Hwang\thanks{Corresponding author}}
\institute{Yonsei University \\
{\tt\small \{juejehun, starforest, seongjae\}@yonsei.ac.kr}}

\maketitle % typeset the header of the contribution

% complex nature image로 바꾸기.,research 안쓰기
%%이 문장에 visual stimuli가 natural 이미지 란 것과 우리가 그것을 recon한다는 내용을 담음. + visual stimuli가 무엇인지까지
% This effort aims to delve deeper into the mechanisms behind human visual information processing. %그리고 이 reconstruction task가 human brain의 작동방식을 이해하는 것 이라는 말로 쓰고 싶었는데 빼는게 맞는듯.
%level과 모달 잇기 이거 생각해야함.
\begin{abstract} 

Understanding how humans process visual information is one of the crucial steps for unraveling the underlying mechanism of brain activity. 
Recently, this curiosity has motivated the fMRI-to-image reconstruction task; given the fMRI data from visual stimuli, it aims to reconstruct the corresponding visual stimuli. 
Surprisingly, leveraging powerful generative models such as the Latent Diffusion Model (LDM) has shown promising results in reconstructing complex visual stimuli such as high-resolution natural images from vision datasets. 
Despite the impressive structural fidelity of these reconstructions, they often lack details of small objects, ambiguous shapes, and semantic nuances. 
Consequently, the incorporation of additional semantic knowledge, beyond mere visuals, becomes imperative.
In light of this, we exploit how modern LDMs effectively incorporate multi-modal guidance (text guidance, visual guidance, and image layout) for structurally and semantically plausible image generations. 
Specifically, inspired by the two-streams hypothesis suggesting that perceptual and semantic information are processed in different brain regions, our framework, Brain-Streams, maps fMRI signals from these brain regions to appropriate embeddings. 
That is, by extracting textual guidance from semantic information regions and visual guidance from perceptual information regions, Brain-Streams provides accurate multi-modal guidance to LDMs. 
We validate the reconstruction ability of Brain-Streams both quantitatively and qualitatively on a real fMRI dataset comprising natural image stimuli and fMRI data.
\end{abstract}

\section{Introduction}
% 이 task에 대한 동기부여 + 이 테스크가 왜 필요한지 제시하기.
% 이전 work들은 어떻게 이 테스크를 수행해 왔는지 제시하기.
%글의 흐름 1. 우리는 visual stimuli reconstruction을 할거다. 2. 우리 또한 다른 work들 처럼 LDM based모델을 사용할거다. 3.그런데 기존 LDM based모델은 high-level semantic이 조금 부족하다고 생각됨. 4. 우리는 3가지 level의 guidance를 LDM모델에 넣어주면 성능이 올라간다는 사실을 발견함. 5. 그래서 이 3가지 level의 guidance를 뇌로부터 효과적으로 추출하기 위해 two-streams hypothesis에 영감을 받음. 6. two-streams hypothesis를 설명. 7. VD에 어떻게 저 3레벨의 guidance를 줄 수 있는지 설명.
%초기 연구는 이라는 말이 들어가야할듯
\begin{figure}[h]
    \centering
    \includegraphics[width=\textwidth]{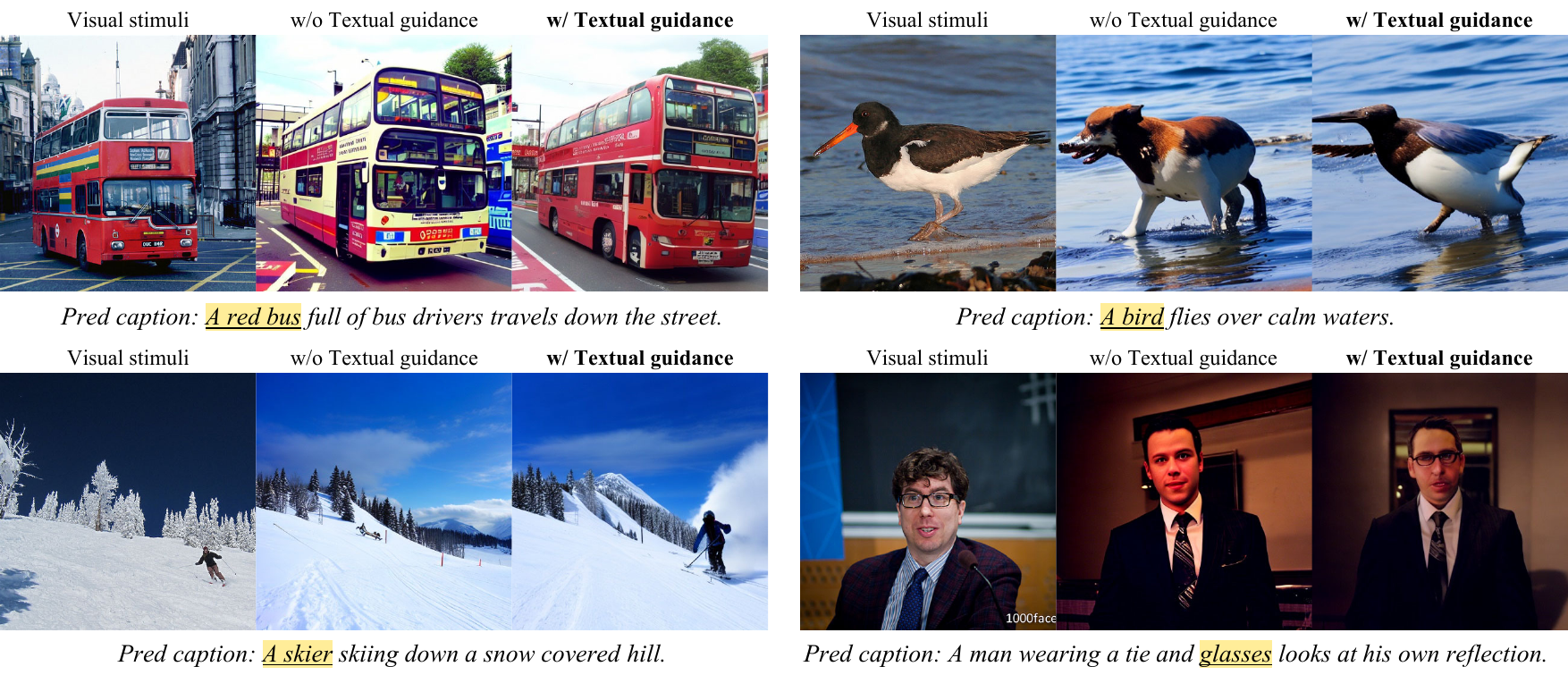} % Set the width of the image to the linewidth
    \caption{Comparison of results demonstrating the impact of textual guidance on visual stimuli reconstruction. 
    In each triplet, the left column displays the original visual stimuli, while the middle and right columns present the reconstructed images without and with textual guidance, respectively. The predicted caption is generated using fMRI data.
    Notably, textual guidance enhances the capture of accurate semantic details, such as glasses and the shape of a bird.}
    \label{fig:comparison_w_imageonly} % Label for referencing with \ref{fig:my_label}
\end{figure}
The human brain's ability to process and interpret visual information is a fundamental aspect of interaction with the world. Efforts to decode this complex process involve research aimed at reconstructing visual stimuli from fMRI data, obtained by exposing subjects to natural images. While simple image reconstructions achieve satisfactory results with  basic models, such as linear mapping~\cite{intro_visual}, reconstructing complex natural images demands precise layout and accurate semantic details, challenging traditional fMRI-to-image mapping methods.

% 자세히 보니 나쁘지 않음
% The human brain's ability to process and interpret visual information is a fundamental aspect of our interaction with the world. 
% In an effort to decode this complex process, research is being conducted to reconstruct visual stimuli from fMRI data, obtained by exposing subjects to natural images.
% Simple stimuli reconstruction achieves satisfactory results with basic models such as linear mapping~\cite{intro_visual}, but reconstructing complex natural images demands precise layout and accurate semantic details, challenging traditional fMRI-to-image mapping methods.

To address these challenges, there has been a shift towards leveraging expressive pretrained generative models, which possess the ability to perceive complex image details.
Starting from early GAN-based models~\cite{mindreader,gu,stylegan2,icgan}, more recent studies using the Latent Diffusion Model (LDM)~\cite{ldm}, known for its training on large datasets and multi-modal capabilities that allow processing of various data types, have proven to be well-suited for handling natural images.
For instance, Takagi et al.~\cite{takagi} reconstructed visual stimuli using Stable Diffusion (SD)~\cite{stablediffusion} by fitting fMRI data to the elements of SD.
Brain-Diffuser~\cite{ozcelik} maps fMRI data to VD-VAE~\cite{vdvae} to generate an initial low-level layout image, followed by reconstruction with Versatile Diffusion (VD)~\cite{versatilediffusion} using CLIP~\cite{clip} embeddings.
MindEye~\cite{mindeye} utilizes fMRI data and SD to produce low-level images, which are then conditioned by image embeddings generated from a diffusion prior, culminating in the final reconstruction with the img2img~\cite{img2img} technique through VD.

Nonetheless, while the above approaches yield promising results, they fall short in accurately capturing semantic details, essential for the precise identification and understanding of specific objects within visual stimuli.
This issue can be observed in Fig.~\ref{fig:comparison_w_imageonly}.
In the center image of each triplet, crucial semantic details are absent (e.g., missing glasses in the bottom right image). 
However, the accurate restoration of semantic details is enabled in the rightmost image of each triplet, where precise textual guidance is provided through predicted captions.
Therefore, to supplement missing semantic details in reconstructed images, our strategy involves providing multi-modal guidance, including precise textual guidance, to the LDM.
Building on this approach, we employ a method that offers the LDM three levels of multi-modal guidance: high-, mid-, and low-level.
High-level guidance introduces accurate semantic details (e.g., the class of entities and the presence of objects) through predicting captions, while the low-level focuses on the basic image layout.
The mid-level guidance incorporates both the rough semantic information and the perceptual features.
A key part of our pipeline is providing precise textual guidance, which correctly includes semantic information about visual stimuli.
To achieve this, our model generates captions using fMRI data and then refines these generated captions through a large language model (LLM). This process of creating precise textual guidance, combined with the two levels of multi-modal guidance ensure that VD receives perceptual and semantic details for visual stimuli reconstruction.

To derive these three levels of guidance from fMRI, we are inspired by \textit{the two-streams hypothesis}~\cite{twostream1,twostream2,twostream3,twostream4}, which suggests that these levels of guidance can be individually extracted from specific brain regions.
\textbf{First}, the \texttt{ventral} visual cortex processes semantic information, such as the existence of objects and their classes.
\textbf{Second}, the \texttt{early} visual cortex, contains perceptual information related to the overall image, which is associated with the low-level aspects of the image.
\textbf{Third}, the \texttt{nsdgeneral} region, which covers regions incorporating aspects of both \texttt{ventral} and \texttt{early} visual cortex partially, containing comprehensive semantic and visual information.
Utilizing a brain region-specific approach, we efficiently extract semantic and perceptual information from fMRI, achieving highly accurate visual stimuli reconstruction.

\noindent
\textbf{Our contributions:}
\textbf{(1)} We propose a new fMRI-to-image reconstruction, Brain-Streams, that extracts three levels of guidance (high, mid, and low) from specific regions of the brain to offer multi-modal guidance to VD.
\textbf{(2)} We have made it possible to reconstruct not only the visual stimuli but also the corresponding captions refined by LLM providing detailed semantic information to VD.
\textbf{(3)} By employing the above method, we achieved state-of-the-art (SOTA) performance on the NSD dataset~\cite{nsddataset} for the visual stimuli reconstruction.

\section{Methodology: Brain-Streams} 
Here we present Brain-Streams, a new pipeline for visual stimuli reconstruction leveraging the two-streams hypothesis with multi-modal guidance.
To provide both textual and visual guidance simultaneously, we utilize Versatile Diffusion (VD), which is capable of integrating these modalities into the reconstruction process.
For the high-level, aimed at delivering detailed semantic information to VD, we utilize fMRI data extracted from the \texttt{ventral} visual cortex, where semantic information is stored. 
The low-level ensures the overall layout of the reconstructed images by extracting perceptual information from the \texttt{early} visual cortex.
The mid-level provides VD with rough semantic insights by utilizing the \texttt{nsdgeneral} region, which encompasses features of both the \texttt{ventral} area and the \texttt{early} visual cortex.
Through Brain-Streams, VD reconstructs visual stimuli, guided by a combination of high-level textual guidance and mid-level visual guidance, grounded on the foundational layout from the low-level.
\begin{figure}[t]
    \centering
    \includegraphics[width=\textwidth]{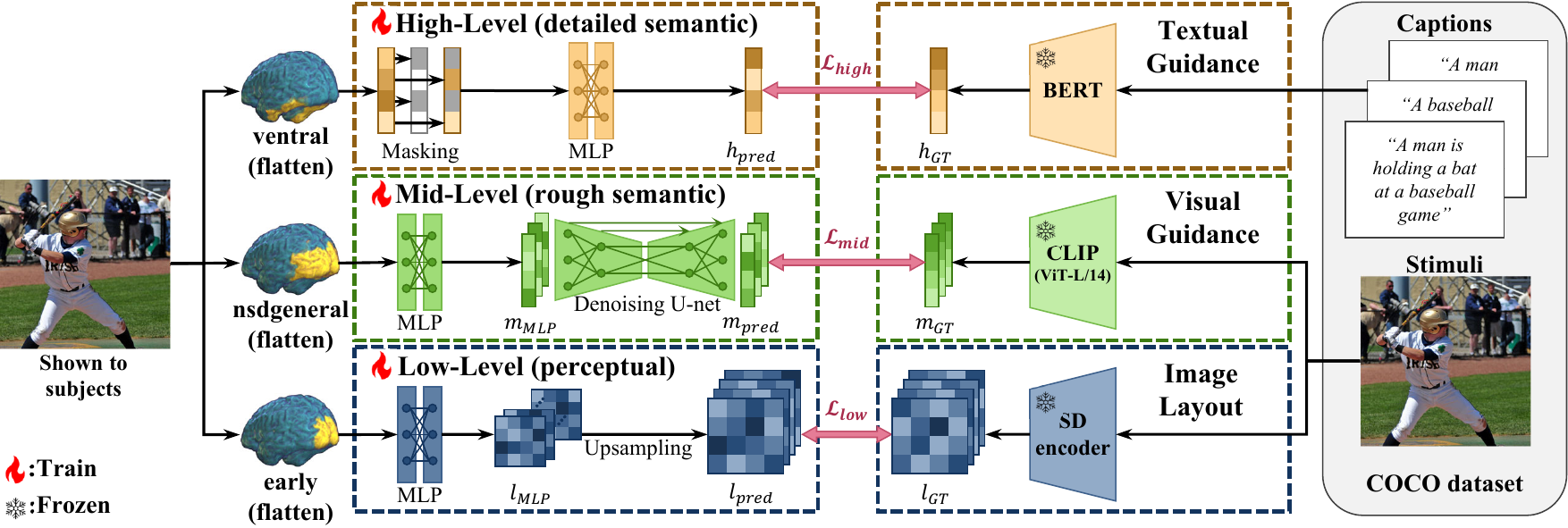} % Set the width of the image to the linewidth
    \caption{Illustration of training pipeline. 
    High-level focuses on the mapping between the fMRI data and BERT's latent vectors to make textual guidance. 
    In the mid-level, we map the fMRI data to the distribution of CLIP image embeddings, which will serve as rough semantic visual guidance.
    For low-level, training progresses by aligning SD's latent vectors with the fMRI data to provide perceptual guidance.}
    \label{fig:Training pipeline}
\end{figure}
\subsection{Training pipeline}
Fig.~\ref{fig:Training pipeline} illustrates our training pipeline.
The models in the left box are learnable, while the models in the right box are kept frozen. 
All fMRI data are flattened.
Models at each level are trained with independent loss functions.

\noindent
\textbf{(\uppercase\expandafter{\romannumeral1}) High-level textual semantic guidance.}
To obtain detailed semantic information, we utilize captions from the MS COCO dataset~\cite{cocodataset}. 
There are five captions corresponding to a single image.
These captions are descriptions written by different individuals after viewing the same images. 
We insert these captions into Optimus-BERT~\cite{optimus,bert}, used in the VD, to generate ground-truth latent vectors $h_{GT}\in\mathbb{R}^{768}$. 
These vectors are predicted using \texttt{ventral} visual cortex fMRI data. 
During training, to construct $h_{GT}$, one of five captions is randomly selected.
Subsequently, we employ an MLP backbone with a masking technique on the fMRI data, obscuring specific information to enhance the model's robustness.
This technique mirrors the masking method used in BERT, applying a similar approach to mask parts of the input in the MLP backbone.
The objective function $\mathcal{L}_{\text{high}}$ is MSE loss between $h_{GT}$ and $h_{pred}\in\mathbb{R}^{768}$. 

\noindent
\textbf{(\uppercase\expandafter{\romannumeral2}) Mid-level visual semantic guidance.}
The CLIP image embedding, utilized as the visual guidance for VD, provides the overall rough semantic information of the generated image. 
The objective of mid-level training is to map fMRI data to this image embedding.
To obtain the ground-truth image embedding, we interpolate the given visual stimuli to a size of 512x512 and then input it into CLIP-ViT-L/14.
To predict this from \texttt{nsdgeneral} region fMRI data, we pass it through an MLP backbone to obtain a predicted embedding of the same size as the CLIP image embedding: $m_{MLP}, m_{pred}, m_{GT}\in\mathbb{R}^{257\times768}$.
Subsequently, we design a DDPM~\cite{ddpm} from scratch to produce realistic image embedding. 
The objective function diverges from the $\epsilon$-prediction approach utilized by Ho et al.~\cite{ddpm}. 
Instead, we calculate the DDPM loss to directly predict the unnoised image embedding, similar to the method described in~\cite{dalle2}. 
Also to ensure a minimum quality threshold for the overall image embedding, we combine Huber loss with the DDPM loss. 
Furthermore, to avoid overly precise fitting to similar images present in the training dataset, we apply batch-wise NCE loss, thereby adjusting the training process to enhance the cosine similarity among similar images.
Then the total loss for mid level is $\mathcal{L}_{\text{mid}}= \gamma_{\text{DDPM}}\mathcal{L}_{\text{DDPM}} + \gamma_{\text{Huber}}\mathcal{L}_{\text{Huber}} +\gamma_{\text{nce}}\mathcal{L}_{\text{nce}}$, where each $\gamma$ are the coefficients of each losses.

\noindent
\textbf{(\uppercase\expandafter{\romannumeral3}) Low-level perceptual guidance.} 
To obtain the low-level layout images, we employ Stable Diffusion (SD). 
Our approach is to map the fMRI data from the \texttt{early} visual cortex to the latent vector of SD's encoder via an MLP backbone and a CNN decoder.
After passing through this MLP backbone, the output is $l_{MLP}\in\mathbb{R}^{16\times16\times64}$. 
$l_{MLP}$ is then upscaled to $l_{pred} \in \mathbb{R}^{64\times64\times4}$ using a CNN decoder, aligning with the latent vector dimensions of SD, and training employs the Huber loss to compare $l_{pred}$ with $l_{GT}$. 
The use of Huber loss is intended to leverage the strengths of both L1 and L2 losses.
To improve low-level performance, we enhance our approach by distilling the knowledge of VICRegL~\cite{vicregl} with $\alpha$ = 0.75 implemented in a ConvNext-XXL architecture, following the methodology used in mindeye~\cite{mindeye}. 
The overall loss for low-level is $\mathcal{L}_{\text{low}}= \gamma_{\text{Huber}}\mathcal{L}_{\text{Huber}} +\gamma_{\text{aux}}\mathcal{L}_{\text{aux}}$, where each $\gamma$ are the coefficients of corresponding losses.

\begin{figure}[t]
    \centering
    \includegraphics[width=\textwidth]{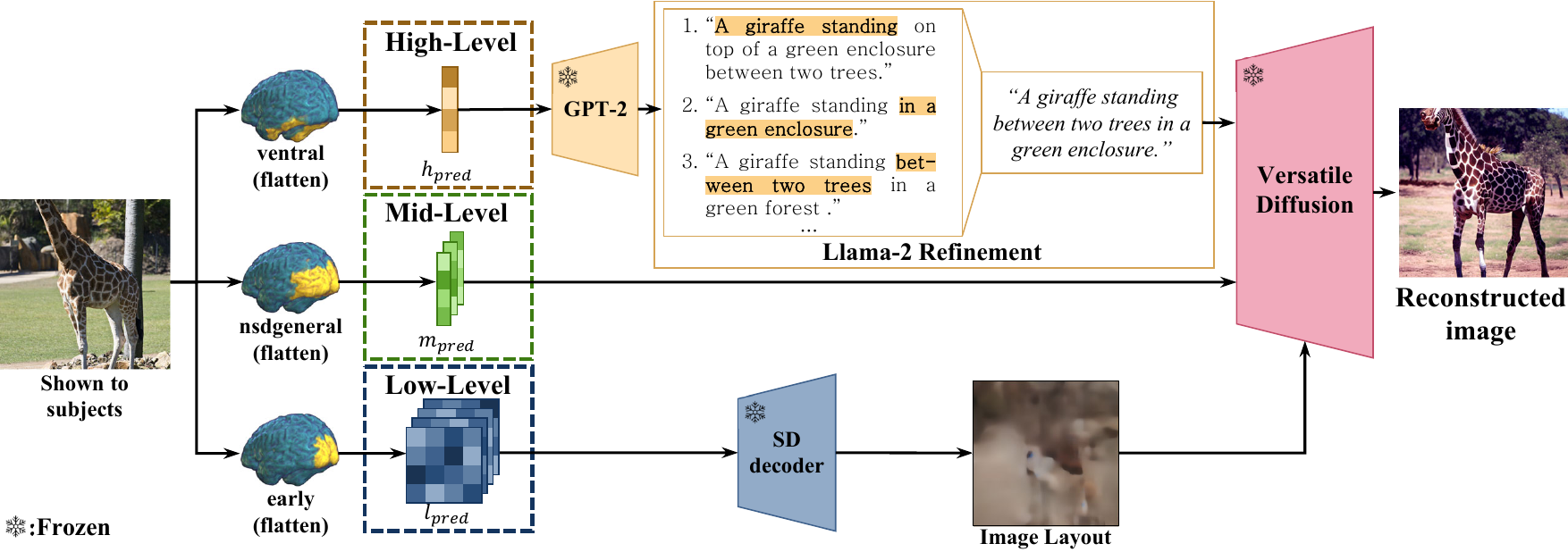} % Set the width of the image to the linewidth
    \caption{Illustration of inference pipeline. 
    In high-level, $h_{pred}$ is predicted from \texttt{ventral} region and decoded with GPT-2. Then Llama-2 refines them to make one cohesive sentence. 
    Image embeddings, serving as visual semantic guidance for VD, are predicted from the \texttt{nsdgeneral} region at the mid-level.
    At the low-level, $l_{pred}$ predicted from \texttt{early} visual cortex is decoded with SD decoder to generate a image layout. 
    This image layout, along with refined caption and predicted image embedding, is then fed into the VD using an img2img approach to produce the final reconstructed image.}
    \label{Testing pipeline} % Label for referencing with \ref{fig:my_label}
\end{figure}
\subsection{Inference pipeline}
Fig.~\ref{Testing pipeline} illustrates the inference pipeline. 
Visual stimuli reconstruction is conducted by utilizing three multi-modal guidance from the trained models.
At high-level, $h_{pred}$ is decoded by GPT-2~\cite{gpt2}.
Due to the nature of GPT-2 as a VAE, the decoding process presents variation, resulting in multiple interpretations from a single encoded state.
To obtain a natural and accurate caption, we conduct multiple decodings and refine the resulting captions using Llama-2~\cite{llama}, ensuring the synthesis of a cohesive and precise output.
The refined caption serves as the textual guidance for VD, providing accurate semantic details to the reconstructed image.
For the mid-level, $m_{pred}$ acts as a visual semantic guidance for VD. 
At the low level, $l_{pred}$ is used to generate an image layout through SD's decoder. 
Following this, the three levels of multi-modal guidance are integrated into VD using an img2img approach, effectively crafting the final reconstruction image with accurate semantic details.
\begin{figure}[t]
    \centering
    \includegraphics[width=\textwidth]{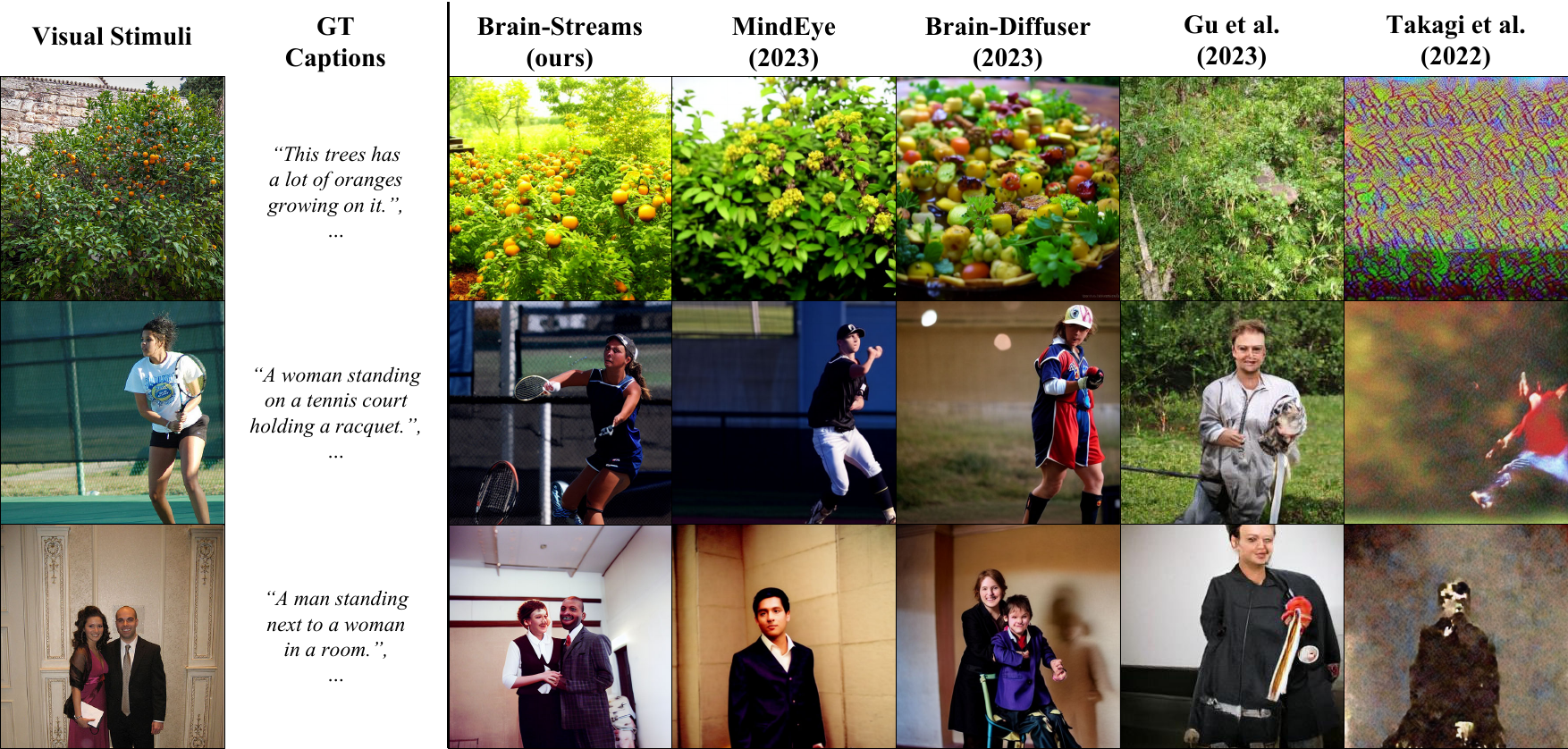} % Set the width of the image to the linewidth
    \caption{Qualitative comparison of reconstructed images with other methods. The first two row are the original visual stimuli and corresponding ground truth captions.}
    \label{fig:comparison_w/_others} % Label for referencing with \ref{fig:my_label}
\end{figure}
\section{Experiments}
All experiments were conducted on a single A6000 GPU.
The implementation details are provided in the supplementary material.

\noindent
\textbf{Dataset and Pre-processing.} We use the NSD~\cite{nsddataset} which provides fMRI data paired with visual stimuli which are part of the MS COCO~\cite{cocodataset} dataset. 
About $30\sim40$ sessions were conducted per subject, but we only trained the models on four subjects (1,2,5,7) who completed all 40 sessions. 
The training dataset comprises 24,980 fMRI-image pairs per subject.
For inference, we employ a test set of 982 images, commonly viewed by all four subjects, which are excluded from the training set. 
In the test set, average fMRI values are derived from presenting the same image to subjects three times and averaging their responses for each image.
We use 1.8mm voxel spaced single-trial betas output from GLMSingle~\cite{GLMsingle} for the fMRI data.
And we also conduct session-wise z-scoring for the fair interpretation of fMRI responses. 
Additionally, we use the streams and nsdgeneral atlas provided by NSD.
\subsection{fMRI-to-image reconstruction results}
\textbf{Qualitative Results.} Fig.~\ref{fig:comparison_w/_others} illustrates a qualitative comparison of our reconstruction results with other methods.
All samples are from subject1.
These results demonstrate that Brain-Streams produces reconstructions that are more faithful to the visual stimuli compared to other methods. 
The superior accuracy in capturing semantic details by Brain-Streams is attributed to the provision of corresponding predicted textual guidance to VD.

\begin{table}[t]
\caption{Quantitative evaluation of our model's reconstruction performance compared to other methods. The values of other methods are from their respective original papers.}
\label{tab:Quantitative}
\centering
\scalebox{0.935}{
\begin{tabular}{lcccccccc}
\toprule
\multicolumn{1}{c}{Method} & \multicolumn{4}{c}{Low-Level (perceptual)} & \multicolumn{4}{c}{High-Level (semantic)} \\
\cmidrule(lr){2-5} \cmidrule(lr){6-9}
 & PixCorr$\uparrow$ & SSIM$\uparrow$ & Alex(2)$\uparrow$ & Alex(5)$\uparrow$ & Incep$\uparrow$ & CLIP$\uparrow$ & Eff$\downarrow$ & SwAV$\downarrow$ \\
\midrule
Lin et al.~\cite{mindreader}       & –       & –       & –       & –       & 78.2\%  & –    & –    & –    \\
Takagi et al.~\cite{takagi}        & –       & –       & 83.0\%  & 83.0\%  & 76.0\%  & 77.0\%& –    & –    \\
Gu et al.~\cite{gu}                & 0.150   & 0.325   & –       & –       & –       & –       & 0.862& 0.465\\
Brain-Diffuser~\cite{ozcelik}      & 0.254   & 0.356   & 94.2\%  & 96.2\%  & 87.2\%  & 91.5\%& 0.775& 0.423\\
MindEye~\cite{mindeye}             & 0.309   & 0.323   & \textbf{94.7\%}  & \textbf{97.8\%}  & 93.8\%  & 94.1\%& \textbf{0.645}& 0.367\\
\rowcolor{gray!25} \textbf{Brain-Streams (ours)} & \textbf{0.342} & \textbf{0.365} & \textbf{94.7\%} & 97.0\% & \textbf{94.0\%} & \textbf{95.2\%} & 0.651 & \textbf{0.357}\\ 
\bottomrule
\end{tabular}}
\end{table}
\noindent
\textbf{Quantitative Results.} Consistent with the conventional evaluation metrics used in previous studies~\cite{ozcelik,mindeye}, Table~\ref{tab:Quantitative} presents a comparative analysis of our model's reconstruction performance against other methods.
The evaluation includes pixelwise correlation (PixCorr) between ground truth and reconstructed images, the structural similarity index metric (SSIM), and two-way identification percentages from several neural network layers, such as the second and fifth layers of AlexNet~\cite{alex}, the last pooling layer of InceptionV3~\cite{incep}, and the output layer of CLIP-ViT-L/14. 
Additionally, distance metrics from EfficientNet-B1 (“Eff”)~\cite{eff} and SwAV-ResNet50 (“SwAV”)~\cite{swav} compare the original images with their reconstructed counterparts.
Upward arrows signify preference for higher values, downward arrows for lower, with bold numbers indicating top performance.
Our values are average across the four subjects, consistent with other methods.
Brain-Streams achieves SOTA performance in almost all metrics, with emphasis on CLIP as a critical measure for accurate semantic details.
This outcome underscores the effectiveness of our method in capturing semantic details. 

\subsection{Ablation studies}
\begin{table}[t]
\caption{Comparing reconstruction results via the usage of each guidance levels.} \label{tab:each level guidance}
\centering
\setlength{\belowrulesep}{0pt}
\setlength{\aboverulesep}{0pt}
\begin{tabular}{lcccccccc}
\toprule
\multicolumn{1}{c}{Guidance} & \multicolumn{4}{c}{Low-Level (perceptual)} & \multicolumn{4}{c}{High-Level (semantic)} \\
\cmidrule(lr){2-5} \cmidrule(lr){6-9}
 & PixCorr$\uparrow$ & SSIM$\uparrow$ & Alex(2)$\uparrow$ & Alex(5)$\uparrow$ & Incep$\uparrow$ & CLIP$\uparrow$ & Eff$\downarrow$ & SwAV$\downarrow$ \\
\midrule
Only High Level       & 0.057       & 0.324       & 69.8\%       & 83.6\%       & 85.9\%    & 86.0\%    & 0.792    & 0.514    \\
Only Mid Level     & 0.180   & 0.297   & 86.5\%  & 95.6\%  & 93.9\%  & 94.3\% & 0.655 & 0.371\\
Only Low Level     & 0.419   & 0.526   & 85.6\%  & 87.4\%  & 65.9\%  & 63.4\% & 0.964& 0.665\\
Mid-High Level   & 0.152       & 0.325      & 86.1\%       & 94.9\%       & 93.9\%    & 94.2\%    & 0.657    & 0.373    \\
Low-High Level       & 0.348       & 0.394       & 90.3\%       & 91.9\%       & 88.0\%    & 86.5\%    & 0.770    & 0.463    \\
Low-Mid Level      & 0.319   & 0.329   & 93.9\%  & \textbf{97.2\%}  & 93.8\%  & 94.2\% & \textbf{0.651}& 0.360\\
\textbf{Brain-Streams} & \textbf{0.342} & \textbf{0.365} & \textbf{94.7\%} & 97.0\% & \textbf{94.0\%} & \textbf{95.2\%} & \textbf{0.651} & \textbf{0.357}\\ 
\bottomrule
\end{tabular}
\end{table}

\begin{table}[t]
\caption{Comparative analysis of Llama-2 refinement over varying decoding samples.} \label{tab:decode sample number}
\centering
\setlength{\belowrulesep}{0pt}
\setlength{\aboverulesep}{0pt}
\begin{tabular}{lcccccccc}
\toprule
\multicolumn{1}{c}{Method} & \multicolumn{4}{c}{Low-Level (perceptual)} & \multicolumn{4}{c}{High-Level (semantic)} \\
\cmidrule(lr){2-5} \cmidrule(lr){6-9}
 & PixCorr$\uparrow$ & SSIM$\uparrow$ & Alex(2)$\uparrow$ & Alex(5)$\uparrow$ & Incep$\uparrow$ & CLIP$\uparrow$ & Eff$\downarrow$ & SwAV$\downarrow$ \\
\midrule
w/o Llama       & 0.338       & 0.362       & 94.2\%       & 96.7\%      & 93.8\%    & 94.3\%    & 0.654    & 0.358    \\
w/ Llama (5s)       & 0.339       & 0.362       & 94.2\%       & 96.6\%       & 93.6\%    & 94.3\%    & 0.653    & 0.357   \\
w/ Llama (15s)  & \textbf{0.342} & \textbf{0.365} & \textbf{94.7\%} & \textbf{97.0\%} & \textbf{94.0\%} & \textbf{95.2\%} & 0.651 & 0.357\\ 
w/ Llama (30s)       & 0.338       & 0.362       & 94.4\%       & 96.6\%       & \textbf{94.0\%}    & \textbf{95.2\%}    & \textbf{0.650}    & \textbf{0.356}    \\
\bottomrule
\end{tabular}
\end{table}

\noindent
\textbf{Effect of Guidance.} We conducted a quantitative evaluation based on the usage of each guidance level.
As seen in Table~\ref{tab:each level guidance}, high-level guidance is associated with higher scores in semantic metrics, whereas low-level guidance achieves higher scores in perceptual metrics. 
Mid-level guidance reflects a balance, encompassing results from both aspects. 
The lower performance observed with exclusive high-level guidance can be attributed to its focus on conveying detailed semantic information without a broader context, resulting in lower scores. 
However, when foundational overall information is supplemented with the additional detailed guidance that high-level inputs provide, it can lead to outstanding outcomes, as demonstrated by Brain-Streams.
Furthermore, Fig.~\ref{fig:comparison_w_imageonly} illustrates the differences between the Low-Mid Level and Brain-Streams methods. 
These disparities have contributed to the divergence in their quantitative measures. 
Further visual distinctions can be found in the supplementary materials.
% As anticipated, employing all levels of guidance yields the best performance, both perceptually and semantically, with particularly high scores in the CLIP metric, supporting our claim.

\noindent
\textbf{Effect of Llama-2 refinement.} Table~\ref{tab:decode sample number} illustrates the impact of Llama-2 refinement on the decoded captions, varying by the number of samples. 
The first row shows results from a single GPT-2 decoded sample without Llama-2 refinement. 
Subsequent rows display outcomes after Llama-2 refinement for 5, 15, and 30 samples. 
The improvement from Llama-2 refinement is evident with an increasing number of samples, with 15 samples being optimal and selected.
\section{Conclusions}
Incorporating multi-modal guidance, especially precise textual guidance, into LDM, our research enhances the visual stimuli reconstruction. 
This advancement marks significant improvements in both perceptual and semantic outcomes. 
Leveraging the two-streams hypothesis, we strategically guide the reconstruction process using data specific to brain regions, addressing challenges in representing semantic details and improving visual stimuli reconstruction. 
We believe these results further imply the validity of the two-streams hypothesis.
% By embedding multi-modal guidance with a focus on precise textual guidance into the LDM process, our research enhances the reconstruction of visual stimuli from fMRI data, marking a notable improvement in both perceptual and semantic outcomes.
% Leveraging the two-streams hypothesis, we strategically guide the reconstruction process with brain region-specific data.
% This approach not only addresses previous challenges in semantic detail representation but also enhances the reconstruction of visual stimuli.
% Our study enhances the reconstruction of visual stimuli from fMRI data by incorporating precise textual guidance into LDM, significantly improving both the visual and semantic accuracy of reconstructed images.

\subsubsection{\ackname}
\clearpage

% \subsubsection{\discintname}

\bibliographystyle{splncs04}
\bibliography{miccai2024}
\end{document}

%fMRI-to-Image Generation with Multi-modal Semantic Guidance. 
%fMRI: fMRI-to-image reconstruction with Multi-modal Representation Input
% \author{Anonymous}

% \institute{Anonymous Organization \\
% \email{**@******.***}}

\begin{table}[ht]
\caption{Implementation details and human rating scores for 11 paired image samples from each method, where 33 participants rated each image's similarity to the original on a scale from 1 to 5.}
\label{tab:hyperparameters_and_versions}
\centering
\scalebox{0.87}{
\begin{tabular}{|c|c|c|c|}
\hline
Hyperparameters & High-Level & Mid-Level & Low-Level \\
\hline
Learning Rate & 0.001 & 0.0005 & 0.001 \\
Batch Size & 8 & 40 & 40 \\
Epochs & 100 & 180 & 170 \\
Optimizer & AdamW & AdamW & AdamW \\
lr scheduler & oneCycle & oneCycle & oneCycle \\
$\gamma$ & - & DDPM: 30, huber: 0.01, nce: 1 & huber: 5.48, aux: 0.1 \\
timesteps & - & 100 & - \\
\hline
img2img strength & \multicolumn{3}{c|}{0.85} \\
text to image ratio & \multicolumn{3}{c|}{0.5} \\
\hline
Model Name & \multicolumn{3}{c|}{Version} \\
\hline
Versatile Diffusion & \multicolumn{3}{c|}{Dual-guided image variation model} \\
Stable Diffusion & \multicolumn{3}{c|}{v1.4} \\
Bert \& Gpt-2 & \multicolumn{3}{c|}{Optimus v1.0} \\
Llama-2 & \multicolumn{3}{c|}{13b for chat completion} \\
\hline
\hline
Methods & \multicolumn{3}{c|}{Human ratings$\uparrow$ (avg $\pm$ std)} \\
\hline
Takagi et al. & \multicolumn{3}{c|}{1.34 $\pm$ 0.24} \\
Gu et al. & \multicolumn{3}{c|}{1.93 $\pm$ 0.43} \\
Brain-Diffuser & \multicolumn{3}{c|}{2.71 $\pm$ 0.47} \\
MindEye & \multicolumn{3}{c|}{2.89 $\pm$ 0.47} \\
\rowcolor{gray!25} Brain-Streams (ours) & \multicolumn{3}{c|}{3.91 $\pm$ 0.73} \\
\hline
\end{tabular}}
\end{table}

\begin{figure}[h]
    \centering
    \includegraphics[width=\textwidth]{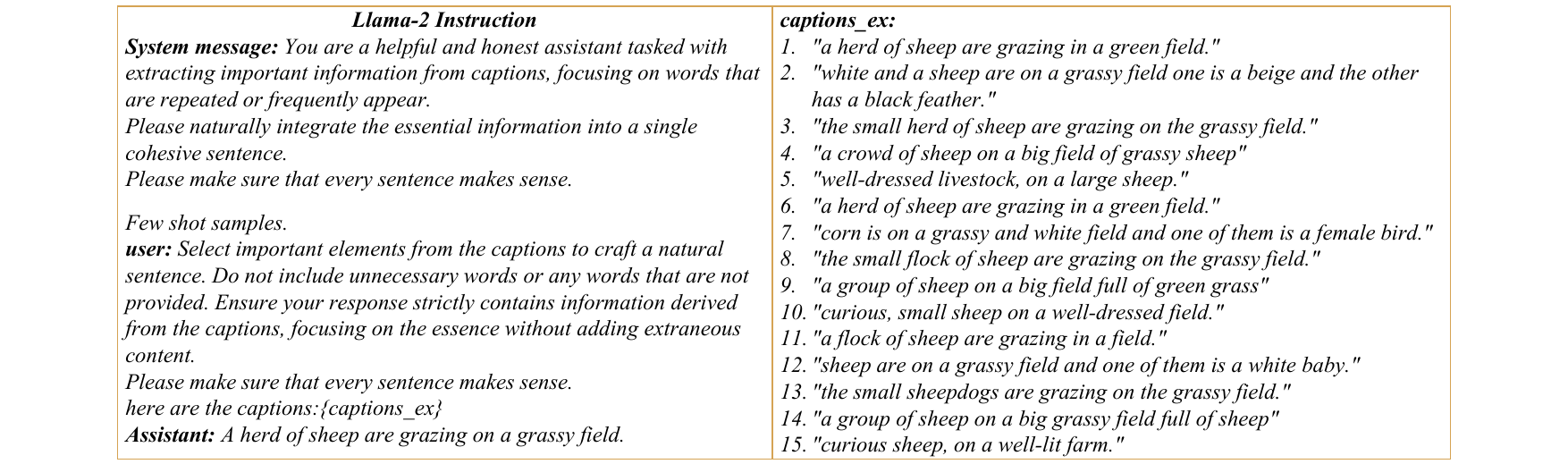} % Set the width of the image to the linewidth
    \caption{Llama-2 refinement instruction.}
    \label{Testing pipeline} % Label for referencing with \ref{fig:my_label}
\end{figure}

\clearpage
\newpage
\begin{figure}[h]
    \centering
    \includegraphics[width=\textwidth]{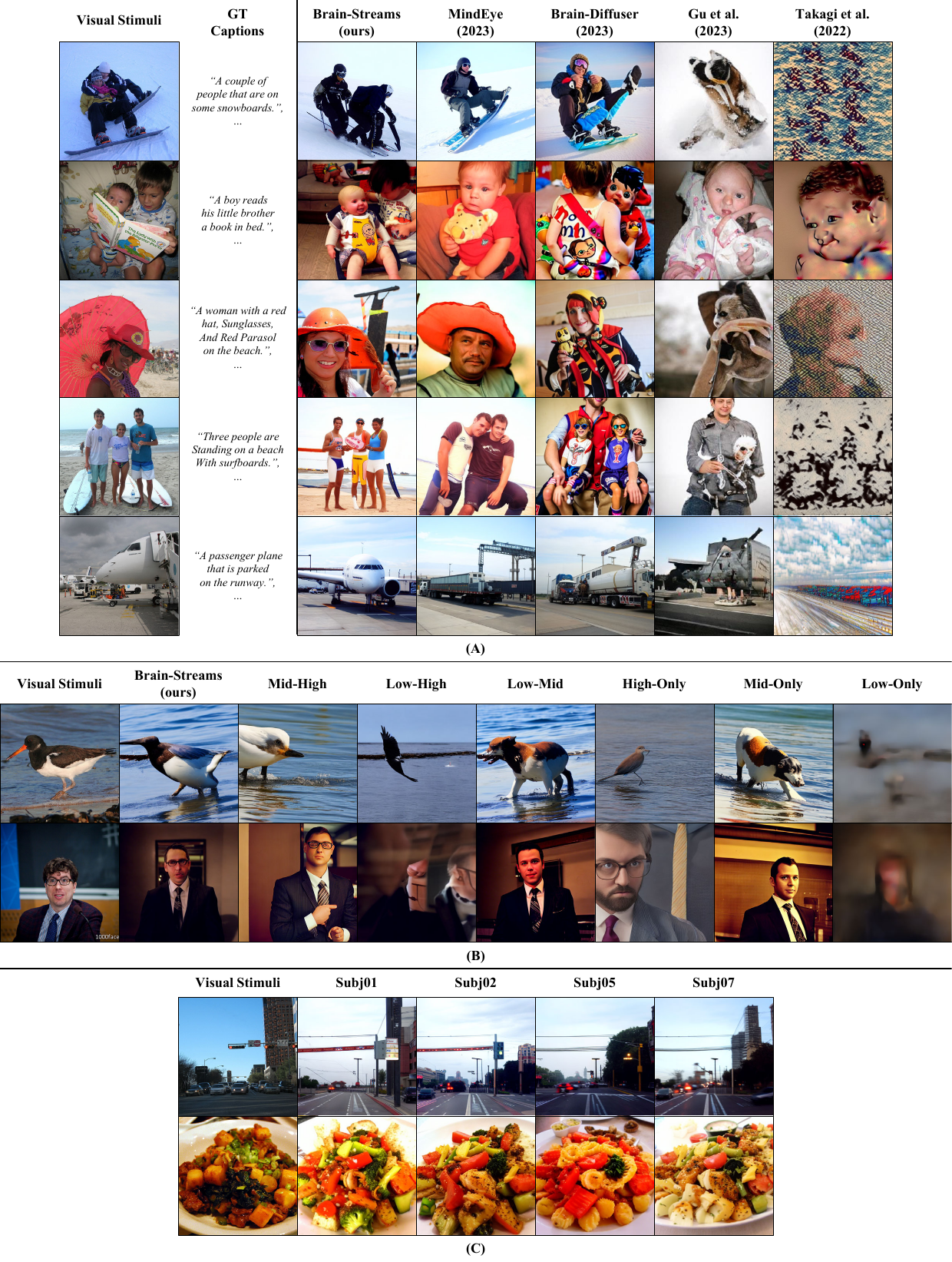} % Set the width of the image to the linewidth
    \caption{(A) illustrates the comparison of our method with others, (B) shows the reconstruction results using each guidance level, and (C) presents the reconstruction results for each subject.}
    \label{supplefigure2} % Label for referencing with \ref{fig:my_label}
\end{figure}